# Data Association Between Perception and V2V Communication Sensors

Mustafa Ridvan Cantas, *Student Member, IEEE,* Arpita Chand, Hao Zhang, Gopi Chandra Surnilla, Levent Guvenc, *Member, IEEE*

*Abstract*—The connectivity between vehicles, infrastructure, and other traffic participants brings a new dimension to automotive safety applications. Soon all the newly produced cars will have Vehicle to Everything (V2X) communication modems alongside the existing Advanced Driver Assistant Systems (ADAS). It is essential to identify the different sensor measurements for the same targets (Data Association) to use connectivity reliably as a safety feature alongside the standard ADAS functionality. Considering the camera is the most common sensor available for ADAS systems, in this paper, we present an experimental implementation of a Mahalanobis distance-based data association algorithm between the camera and the Vehicle to Vehicle (V2V) communication sensors. The implemented algorithm has low computational complexity and the capability of running in real-time. One can use the presented algorithm for sensor fusion algorithms or higher-level decision-making applications in ADAS modules.

## I. Introduction

Many automotive companies invest in connected vehicle applications, such as Left Turn Assist, Intersection Movement Assist, and Collison Avoidance. While these communication-based driver-assist technology applications aim to reduce the number of accidents on the road, some existing driver-assist technologies aim to avoid accidents by detecting the threats using range sensors such as cameras and radar. It is crucial to identify if both systems refer to the same target to warn the driver in a potential collision scenario accurately. This problem is called a data association problem. This manuscript focuses on identifying the association between the two different sensor measurements of a camera (range sensor) and a Vehicle to Vehicle (V2V) communication modem. After the data association step, a higher-level module, i.e., threat assessment module, would use the data association results to take necessary safety precautions.

There are three main categories for the data association problem which are used for tracking and sensor fusion applications. The first one is the measurement to measurement association, which is also called the track initiation problem. The second one is the measurement to track association, which is also called a track maintenance problem. The final one is the track to track association problem, which is also called the track fusion problem. Among these, this paper focuses on the implementation of the track to track data association.

This part will summarize some of the data association work in the literature. Some of the commonly used perception/range sensors in ADAS are camera, lidar, radar, and ultrasonic sensors. Different working principles of these sensors result in different resolutions, ranges, and detection rates depending on the sensor's physical characteristics and environmental conditions. The strength, weaknesses, and working principles of these sensors are presented in [17]. Connected and autonomous vehicles use Vehicle to Everything (V2X) [11] communication as another sensor alongside the existing perception/range sensors, which improves connected vehicles' self-awareness. In autonomous vehicles, Multi-Sensor Multi-Object Tracking Modules handle sensor fusion tasks, which increases the accuracy of detections by combining the measurements from multiple sensors. One essential stage of the Multi-Sensor Multi-Object Tracking is the data association. In [8], the authors developed a Track to Track data association algorithm, which compares the Mahalanobis distance between tracks from multiple sensors. Similarly, in [19], the authors used minimum Mahalanobis distance with the Chi-square test for their application's association problem. In [2,10] Hungarian Algorithm is used for data association. Alongside these methods, there are also some other probabilistic models in the literature. Some of these are Probabilistic Data Association Filter (PDAF), Joint PDAF (JPDAF), and Multi Hypothesis Tracking Algorithm (MHT). In PDAF, instead of using hard assignments, multiple detections inside the validation gate are considered to update the current track [3, 4, 7]. The PDAF is more appropriate for single track scenarios than multi-track scenarios. Therefore, JPDAF is developed for multiple target scenarios [12]. The JPDAF works by calculating the joint probability distribution between the tracks and measurements. Other methods used for multi-object tracking data associations are Multiple Hypothesis Tracking (MHT) [9] and Markov

M. R. Cantas with the Automated Driving Laboratory and the Department of Electrical and Computer Engineering, The Ohio State University, Columbus, OH, 43210 USA. (e-mail: cantas.1@osu.edu)

A. Chand, H. Zhang and G. Surnilla are with the Ford Motor Company, Dearborn, MI 48124 USA. (e-mail: achand6@ford.com, hzhang13@ford.com, gsurnill@ford.com)

L. Guvenc is with the Automated Driving Laboratory, Department of Mechanical and Aerospace Engineering, and Department of Electrical and Computer Engineering, Ohio State University, Columbus, OH 43210 USA (e-mail: guvenc.1 @osu.edu).

Chain Monte Carlo Data Association (MCMCDA) [13, 20] and Markov Decision Process [16]. While all these techniques can be applied for target association, in this paper, the Mahalanobis Distance-based Track to Track association algorithm is preferred because it is not as demanding as the other methods in terms of the computational resources. Also, the Mahalanobis distance is a highly relevant metric that considers the targets' position, measurement uncertainty, and correlations between two tracks [6]. The application area of the data association algorithms can be naturally extended to autonomous vehicles as well [14]. Some of the examples of data association applications in autonomous vehicle research can be seen in [5, 16, 21].

The organization of the rest of the paper is as follows. Section II summarizes the experimental setup and overall architecture with an explanation of the used algorithm. Then the results and conclusion sections conclude the article.

## II. THE EXPERIMENTAL SETUP AND SYSTEM ARCHITECTURE

This section describes the overall architecture of the designed and implemented data association architecture. Our experiment setup is used to depict the general architecture. In our experiment, a host vehicle (HV) and two remote vehicles (RV_1 and RV_2) are used. HV is equipped with a camera and V2V communication onboard unit (OBU), and remote vehicles are fitted with only V2V OBUs. The video stream from the host vehicle and high accuracy RTK GPS measurements for all the vehicles are recorded for ground truth creation purposes. The HV data association system architecture is shown in Fig. 1, which consists of measurements, coordinate transformation (for V2V measurements), synchronization and filtering, buffering, and track to track association modules. When reading the rest of this section, readers can refer to Fig. 1 to see each module's relationship with other modules.

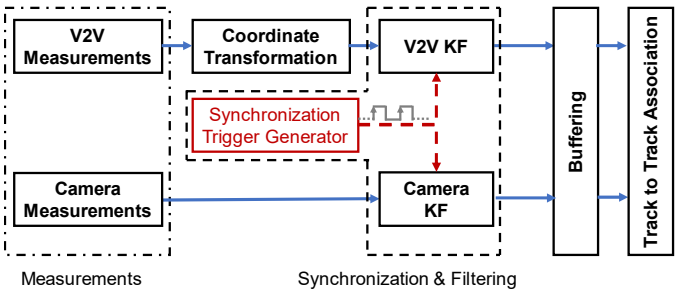

Fig. 1. The designed data association architecture for the HV.

### A. Measurements and Associable Parameters

Most of the new vehicles have a camera sensor as a standard ADAS sensor. Although some of the vehicles also have radar, it is generally available for higher-end vehicles with Adaptive Cruise Control functionality. Therefore, this study focuses on Track to Track Association implementation between the most common ADAS sensor, the camera, and the V2V communication modem. If required, one can quickly adapt the same algorithm to include more than two sensors. In our experimental vehicle architecture, the host vehicle is equipped with an automotive-grade smart, forward-looking camera. The measurement rate of the used camera is around 40 Hz and has a 100-degree field of view. The camera can detect other vehicles on the road and track them. Therefore, the camera has its own object management algorithm, and it publishes the tracked objects' IDs alongside their spatial position, dimensions, type, relative speed, and bearing angle. The host vehicle is also equipped with a V2V OBU, which transmits and receives Basic Safety Messages (BSM) to communicate with other traffic participants. The full definition of BSM is given in [18], which is a message set broadcasted by each connected vehicle at 10Hz. This message set has the ID, global position, heading, speed, dimensions, and path history of the vehicle. Although both sensors offer other measurements, the measurements shown in Table I are considered to be comparable with each other. While this setup is the minimum requirement for the host vehicle, remote vehicles do not necessarily have to have this whole setup. They only need to have V2V OBUs to transmit their BSMs.

TABLE I
SOME OF THE CAMERA AND V2V MEASUREMENT PARAMETERS

| Camera | V2V |
|---|---|
| Id | Id |
| Processor time | UTC Time |
| Lateral Distance | Latitude |
| Longitudinal Distance | Longitude |
| Relative Heading | Heading |
| Relative Speed & Speed | Speed |
| Length | Length |
| Width | Width |

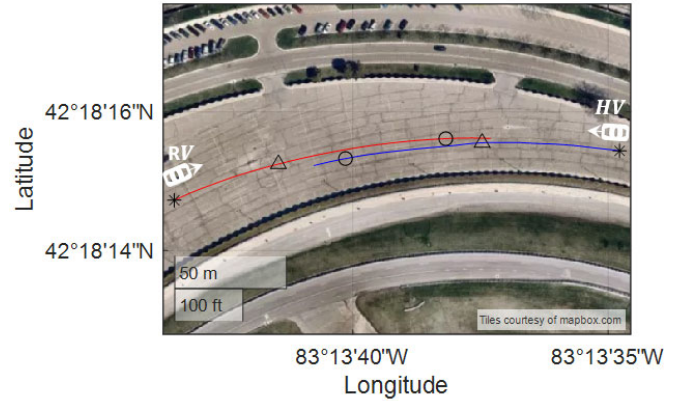

Fig. 2. Paths followed by the host and remote vehicles in the global coordinate system. "*" represents the starting location of the paths. Vehicles start to move from the '*' mark and proceed to 'Δ' and 'o'.

It is essential to determine associable parameters between the two automotive sensors considered here: the camera and the V2V modem, to implement the data association module. This task aims to find common and comparable parameters between these two sensors, which will help identify an ADAS target. In Table I, the relevant measurement parameters for the automotive-grade smart camera and V2V module are listed. We conducted experiments to verify which one of these parameters is associable. In the experiments, a host vehicle and a remote vehicle drive opposite directions in a slightly curved path, which is shown in Fig. 2. Vehicles' starting location and positions with five-second intervals are marked with '*', 'Δ', and 'o' respectively on their trajectories to reflect vehicles' motion throughout time. After having all measurements in the host vehicle coordinate system, measurement parameters are analyzed, which are to be used for the data association task. It is observed from the plots in Fig. 3 that the relative distance/offset measurements from the camera and V2V are comparable to each other. Similarly, relative heading between the Host Vehicle and Remote Vehicle and lateral and longitudinal velocities from camera and V2V are also comparable with each other. Therefore, these parameters are chosen as associable parameters. However, the Remote Vehicle (length and width) size measurement from the camera does not match with actual dimensions published via V2V communication. Therefore, vehicle size measurements are not considered as a reliable parameter for the data association task.

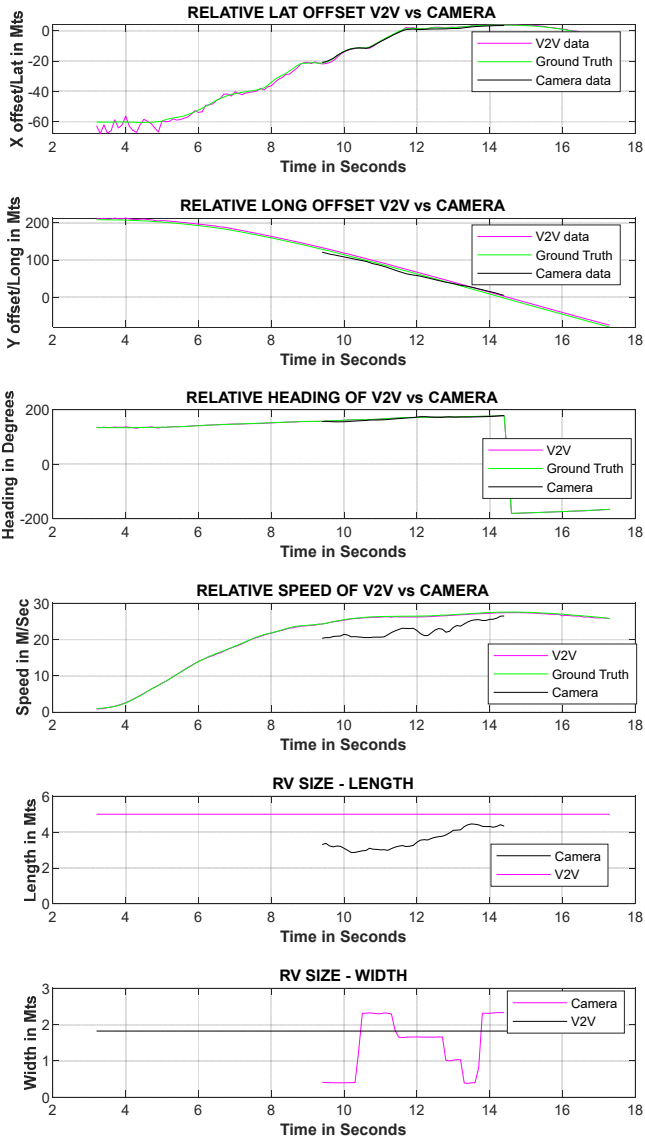

Fig. 3. Camera and V2V sensor measurement performance comparison for data association. Ground truth data is acquired from high precision RTK GPS system.

*B. Coordinate Transformation*

In the constructed architecture, the measurements from the V2V are in the global coordinate system, whereas the measurements from the camera are in the vehicle coordinate system. In Fig. 4, one can see the global and local coordinate frames for the host and remote vehicles. In order to associate measurements between the HV sensors, both measurements must be in a common coordinate system. For this purpose, the V2V measurements representing the remote vehicle positions are transformed into the host vehicle coordinate system. Thus, it becomes possible to compare the camera measurements with the V2V measurements. The mentioned coordinate transformation requires the knowledge of the Host Vehicle (HV) and Remote Vehicle's (RV) global coordinates and heading, which are measured by the V2V modems. In the translation step, the global coordinates (latitude, longitude) of the HV and RV have transformed into Universal Transverse Mercator (UTM) grid coordinates in (1) where deg2utm function is used from [15]. Then the relative distance between the HV and RV is calculated in the North and East direction shown in (2) and (3). In the coordinate transformation step, the calculated relative distances in the global coordinate system are transformed into the HV coordinate system in (4). Finally, the relative distance $d_r$ between the host and remote vehicle can be calculated as in (5).

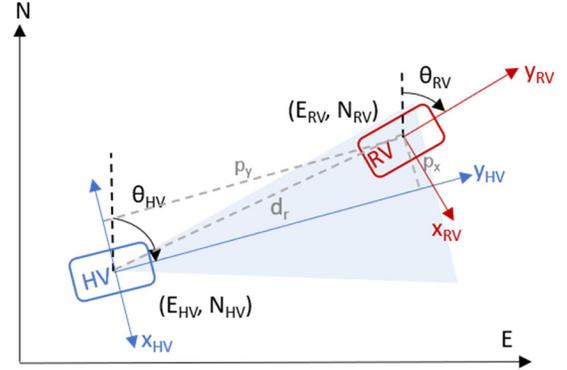

Fig. 4. The coordinate systems for Host Vehicle and Remote Vehicle. Field of view of the camera is represented with blue transparent region.

$$[N, E, utmzone] = \text{deg2utm}(lat, long) \quad (1)$$
$$\Delta E = E_{RV} - E_{HV} \quad (2)$$
$$\Delta N = N_{RV} - N_{HV} \quad (3)$$

$$\begin{bmatrix} p_x \\ p_y \end{bmatrix} = \begin{bmatrix} \cos(\theta_{HV}) & -\sin(\theta_{HV}) \\ \sin(\theta_{HV}) & \cos(\theta_{HV}) \end{bmatrix} \begin{bmatrix} \Delta E \\ \Delta N \end{bmatrix} \quad (4)$$

$$d_r = \sqrt{p_x^2 + p_y^2} \quad (5)$$

*C. Synchronization and Filtering*

For filtering the camera and V2V track measurements, Kalman Filters are used for each sensor type. The purpose of having the filtering stage is to estimate the state of the tracks at the desired time step. It should be noted that V2V position measurements are transferred to the vehicle coordinate system before they are fed to the Kalman Filter. Thus, all the filtering and tracking task is achieved in the vehicle coordinate system. The prediction and update stages of the Kalman Filter Algorithm is given as in Algorithm 1.

---
**Algorithm 1: Kalman Filter**

Prediction:
$$x' = Fx + u$$
$$P' = FPF^T + Q$$

Measurement Update:
$$y = z - Hx'$$
$$S = HP'H^T + R$$
$$K = P'H^T S^{-1}$$
$$x = x' + Ky$$
$$P = (I - KH)P'$$

---

In the Kalman Filter Algorithm the state vector is:
$$x = \begin{pmatrix} p \\ v \end{pmatrix} \qquad (6)$$

where $p$ is position and $v$ is velocity. The state prediction function $F$ *is* the 2x2 matrix in:

$$\begin{pmatrix} p' \\ v' \end{pmatrix} = \begin{pmatrix} 1 & \Delta t \\ 0 & v \end{pmatrix}\begin{pmatrix} p \\ v \end{pmatrix} \qquad (7)$$

The measurement function for camera $H$ *is* the 2x4 matrix in:

$$\begin{pmatrix} p_x \\ p_y \end{pmatrix} = \begin{pmatrix} 1 & 0 & 0 & 0 \\ 0 & 1 & 0 & 0 \end{pmatrix}\begin{pmatrix} p_x \\ p_y \\ v_x \\ v_y \end{pmatrix} \qquad (8)$$

where $p=[p_x \; p_y]'$, $v=[v_x \; v_y]'$ and $Q$ and $R$ represent the covariance matrices for processes noise and measurement noise, respectively.

One should be careful when associating measurements from sensors with different measurement time stamps. Associating measurements without synchronization can result in false data association. Therefore, in our design, data synchronization is done between the camera and V2V OBU sensor measurements. The synchronization is realized periodically with a 10 Hz trigger. Synchronization Trigger Generator in Fig. 1 triggers the camera and V2V Kalman Filters simultaneously at each rising edge of the trigger signal to generate state estimations for each sensor. Thus, two different prediction types are realized in the Kalman Filters (KF). In the first case, if a measurement is received, the KF makes a state prediction and updates the prediction with the received measurement. In the synchronization case, both KFs make another state estimation at the triggered time step, which would be used for the track to track association.

### D. Buffering

In the data association task, if the targets are well separated from each other, it is easier to associate the measurements which are originating from the same target. However, if the targets are getting closer to each other with high spatial uncertainty in the measurement, it is hard to differentiate them from each other. Buffering is introduced to enhance track to track data association performance by creating a track history for each sensor measurement and compare the distance between two tracks over the buffer size. This is especially useful to eliminate false associations due to high spatial uncertainties for short time intervals. An example of where the buffering improves the data association performance is illustrated in Fig. 5. In Fig. 5 the detected vehicles by the camera are positioned almost on top of each other. The ambiguity in the data association process is eliminated by considering the path history of targets. Thus, the correct data association assessment is performed. The buffering formulation is integrated into the track to track data association formula given in (9).

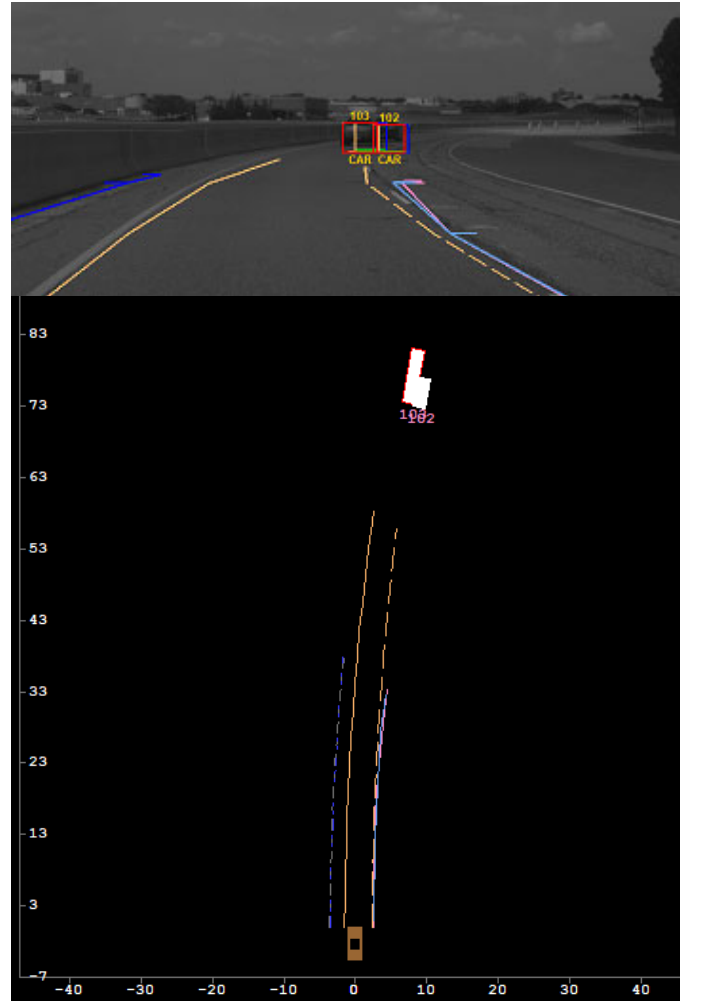

Fig. 5. Sample scenario where the camera measurements for remote vehicles have high spatial uncertainty (Spatial measurements are too close to each other).

### E. Track to Track Data Association

The implemented Track to Track Data Association module in Fig.1 is designed based on the method proposed in [8]. In this module's development, two sensors are considered S1: camera and S2: V2V modem. Each of these sensors outputs a list of detections $A = \{a_1, a_2, \ldots, a_n\}$ and $B = \{b_1, b_2, \ldots, b_m\}$. The designed module finds the detections from each sensor for the same target and associates them. Since the comparison of the detections' location history would improve the data association performance, detections over time are buffered in the Buffering Module to form tracks. Thus, we can define a track as the state estimate vector of a single target over a time interval, where each element of the vector corresponds to a state estimate at the corresponding measurement time. The used algorithm for this task, which uses position similarity for data association, is presented below.

**Algorithm 2: Track-to-Track Association**
1: Concatenate the tracks of all the sensors.
2: Numerate the concatenated tracks from 1 to N.
3: Create an N x N matrix for the Track to Track Distances (TTTD) between the tracks.
    Set the cells over the diagonal entries to infinity ($\infty$) in order to avoid repeating the calculations.
    Set cells that represent the distance between the same sensor measurements to $\infty$.
    Set the remaining cells to Mahalanobis Distance between the corresponding tracks.
    Set the cells to $\infty$, where the distance is larger than a defined threshold.
4: **while** there is any value other than $\infty$ **do**:
  Choose the minimum value in the matrix (row, column).
  **if** corresponding tracks not in a cluster **do**:
    Create a new cluster from these two clusters
  **else if** one of the tracks is in an existing cluster **do**:
    Add the other track into the existing cluster
  **else do**:
    Nothing
  Set the selected column and row cells to infinity for the corresponding sensors.
5: **if** there is a track which is not in any cluster **do**:
  Form new clusters, for tracks that are not in existing clusters.

In the presented algorithm, the Mahalanobis Distance calculation between two tracks $(a, b)$ is performed over the history size of $n$ to measure the position similarity between two tracks as in (9) & (10).

$$D_k^{(a,b)} = \frac{1}{n}\sum_{i=0}^{n-1} d_k^{(a,b)} \quad (9)$$

where

$$d_k^{(a,b)} = \sqrt{(X_k^a - X_k^b)^T (P_k^a + P_k^b)^{-1}(X_k^a - X_k^b)} \quad (10)$$

Here $X_k$ is the state estimate and the $P_k$ is the covariance matrix of the state estimate for the corresponding tracks at time step $k$. The covariance matrices are acquired from the Kalman Filter state estimation for each sensor.

In Fig. 6 and Fig. 7, an example of the matrix representation of the Track to Track Association (TTA) Algorithm stages 3 and 4 are shown. In the shown example, the V2V has two measurements, and the camera has four target detections. In these figures, Tracks are named as $T_{ij}$ where $i$ is the sensor ID, and $j$ is the detection number. For example, $T_{12}$ represents the second V2V detection. Similarly, since the camera is the second sensor $T_{23}$ represents the third detection of the camera sensor.

Fig. 6. Algorithm 2 Stage 3 example for V2V & camera TTA.

Stage 4: Loop

Clusters: {T12, T22}          Clusters: {T12, T22}, {T11, T21}

Fig. 7. Algorithm 2 Stage 4 example for V2V & camera TTA.

## III. RESULTS

Two different experiments are conducted to demonstrate the effectiveness of the implemented algorithm. These two scenarios look at the data association task for car-following and Intersection Movement Assist (IMA) scenarios. Fig. 8 illustrates the travel direction and layout of the vehicles for both scenarios.

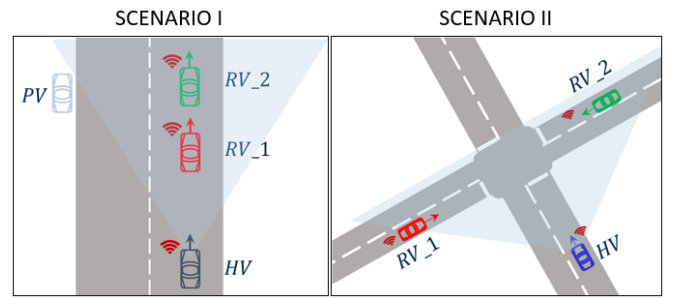

Fig. 8. Left: Car following scenario (Scenario I). Right: IMA scenario (Scenario II).

### A. Scenario I: Car Following

In the first conducted experiment, target vehicles are traveling in front of the host vehicle. Their distance with respect to the host vehicle and to each other changes throughout the experiment Fig. 8 (left). In the figure, RV_1 and RV_2 represent the remote vehicles, PV represents the parked vehicle, and HV represents the host vehicle. Parked vehicles on the track are not transmitting any V2V messages, but the camera can detect them. As for sensors, only the host vehicle has a forward-looking camera capable of detecting the target vehicles and their positions. All three vehicles are equipped with V2V modems. From time to time, remote vehicles travel side by side, but mostly all the vehicles follow each other in the test track shown in Fig. 9, which consists of small curves and straight parts.

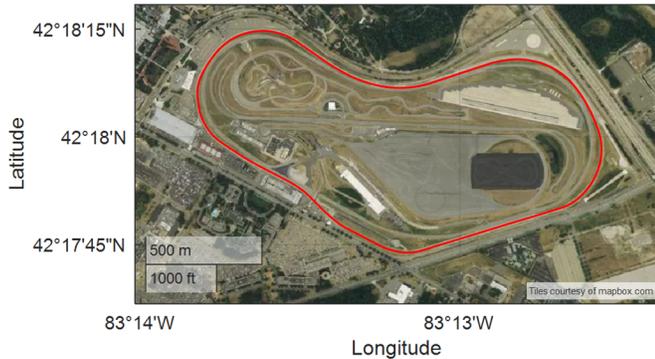

Fig. 9. Test track and recorded trajectory of host vehicle for the car-following scenario (Scenario I).

At each measurement time step, the camera outputs detected objects as a list by identifying each detection with an ID number. Time to time, the camera changes the ID for the same target when an occlusion occurs. V2V BSM messages for remote vehicles are received continuously and they maintain the same IDs throughout the entire experiment. By considering these characteristics of the received measurements, we created ground truth IDs for camera detections of RV_1 and RV_2. It is created by comparing the camera detection IDs with the recorded video stream. In Fig. 10, the upper part represents the RV_1 camera detection IDs, and the middle part represents the RV_2 camera detection IDs over time. The ground truth for the camera IDs corresponding to the remote vehicles is shown with blue color, and the result of the data association is marked with yellow. As can be seen from Fig. 10, most of the data associations are performed successfully. Most of the failures occur for RV_2. This is because the RV_2 is generally partially occluded from the camera ant the accuracy of the camera distance measurement is affected significantly for the occluded objects.

To assess how closely the detected objects are associated, a data association confidence level is introduced (11).

$$Conf = min(0, 100 * \frac{th-D}{th}) \qquad (11)$$

Where $th$ is the selected inter-vehicle distance threshold, and D is the distance between the two tracks. This equation will yield a 100 percent confidence level if the two measurements are exactly the same. On the other hand, if the measurements are separated from each other, it gets closer to 0 confidence. If the two measurements are separated more than the set $th$ distance, the confidence will be 0, showing that selected measurements are not associated. In Fig. 10 it is seen that the confidence level for the remote vehicle RV_1 is between 90-100 percent. On the other hand, for the RV_2, the confidence level varies between 45-100 percent. The larger confidence level variation occurs because the RV_2 is mostly occluded by RV_1, HV's camera could not measure the position of RV_2 accurately.

Track Matching Accuracy (TMA) [19] is used as another performance measure for the TTA association. It is defined as the percentage of correct matching decisions taken by the TTA algorithm among all the test cases. In the conducted experiment, camera measurements were able to be associated with the V2V measurements for RV_1 with 100 percent TMA. However, for the second remote vehicle, TMA drops to 98.8 percent.

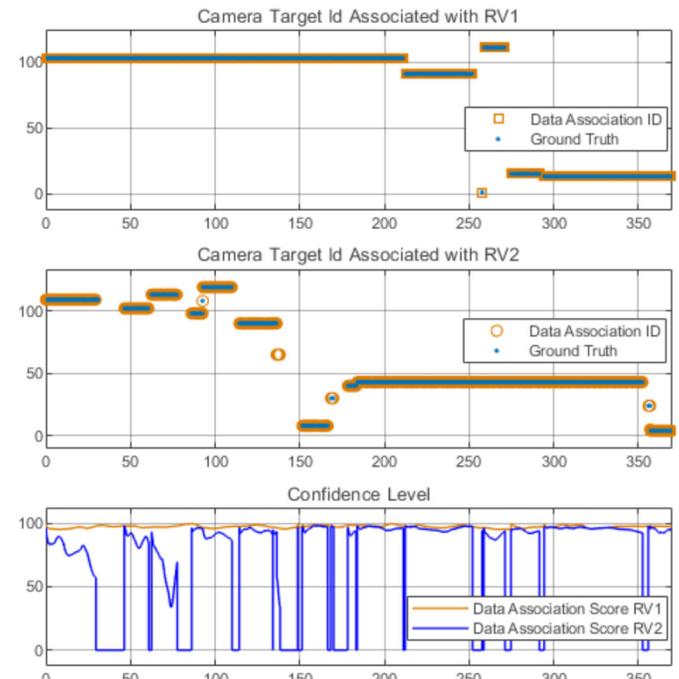

Fig. 10. Scenario 1: Track to Track Association V2V - Camera id assignment for car following scenario.

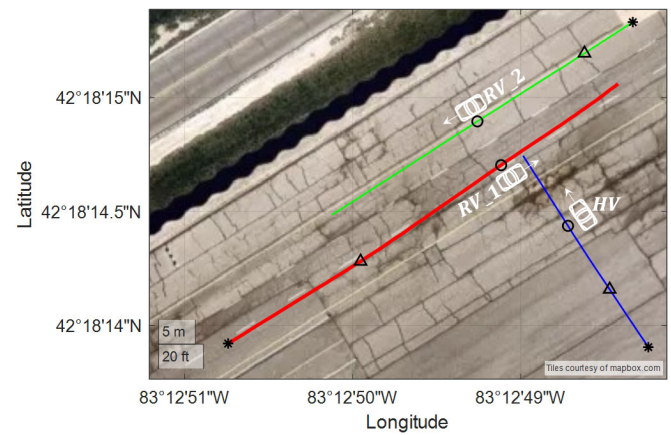

Fig. 11. Recorded trajectories of test vehicles for the IMA scenario (Scenario II). Vehicles' heading and position representations for occlusion instance are shown with car images at the positions marked with 'o'. '*' represents vehicles' starting points and 'Δ' represents vehicles' positions at an intermediate time instance.

## B. Scenario II: Intersection Movement Assist (IMA)

In the second scenario, HV, RV_1 and RV_2 experimental vehicles travel towards an intersection from different directions. In the scenario, the host vehicle aims to match V2V detections with the camera detections to issue an appropriate warning by comparing ADAS outputs from V2V and camera-based systems. In V2V ADAS applications, this scenario corresponds to Intersection Movement Assist (IMA) application. As a V2V safety application, IMA intends to warn the host vehicle driver if there is a high collision chance in an intersection [1]. In the experiment, the HV is moving to the intersection from the southbound of the road. The remote vehicles RV_1 and RV_2 are moving to the intersection from westbound and eastbound, respectively. When the remote vehicles come to the intersection, they continue traveling across each other. On the other hand, the host vehicle stops at the intersection. The layout of the experiment and the vehicles' recorded positions are shown in Fig. 8 (right) and Fig. 11. In Fig. 11, the vehicles' positions in three different time instances are marked on their trajectories to show travel history. Vehicles start from the '*' mark and proceed to 'Δ' and 'o'. At the 'o' mark RV_1 starts to block the RV_2. The camera frame for this instance can be seen in Fig. 13.

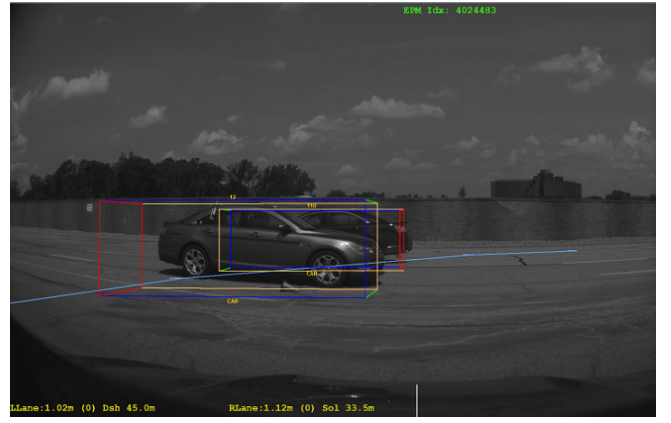

Fig. 13. Top: Captured camera frame just before the RV_1 occludes RV_2.

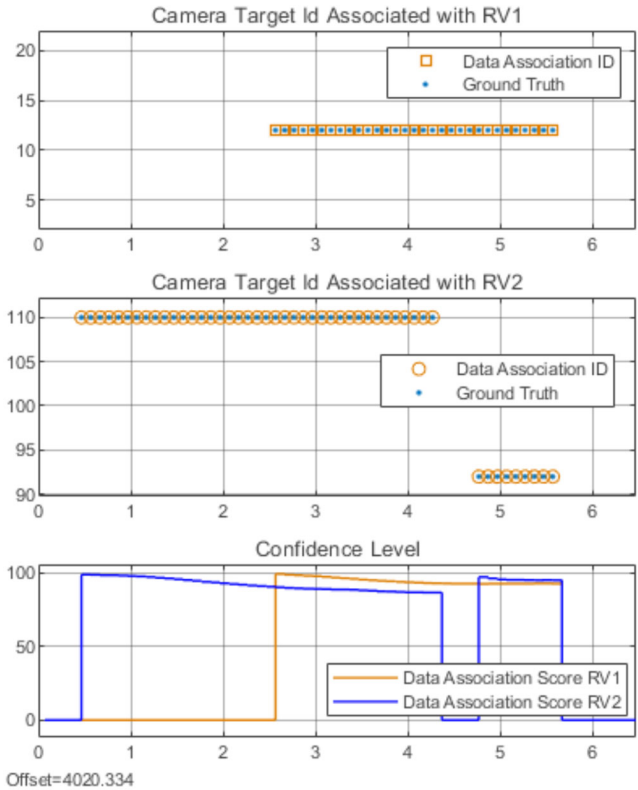

Fig. 12. Scenario 2: Track to Track Association V2V - Camera id assignment for IMA experiment.

In Fig. 12, one can see the V2V camera data association results. While the camera detection ID for the RV_1 is 12 throughout the experiment, RV_2 got two different IDs: 110 and 92. RV_1 blocks the RV_2 for a short time interval (Fig. 13) when the remote vehicles cross each other. After RV_2 becomes visible to the host vehicle, the camera assigns a new ID to RV_2. Data association results in Fig. 12 show that the developed method is implemented successfully. All the data association results match with the ground truth for this scenario.

## IV. CONCLUSION

The designed and implemented data association method aims to identify measurements for the same target from different sensors. Although this method can be generalized and applied to different sensors like radar and lidar, this work focused on the V2V modem and camera measurements. As a first step, associable parameters for the considered sensors were investigated. It was found out that the location, relative heading, and relative speed measurements from these two sensors are comparable, and they can be used for the data association tasks. Among these parameters, the location measurements were the main parameters used in the implemented algorithm. On the other hand, speed and relative heading parameters were also used as another gating to prevent any false data association assessment. This paper contributes to the literature by presenting an experimental implementation of V2V and camera data association method. While the presented method was experimentally tested and shown to be effective for even curved roads and intersections, it is required to conduct more experiments with a higher number of vehicles in more complex scenarios before the deployment of the algorithm. In the future work more advanced algorithms such as JPDA and LSTM based approaches will be considered. As another improvement for the developed method, one can drop the buffering stage for V2V and use the path history information available in the BSM message set.

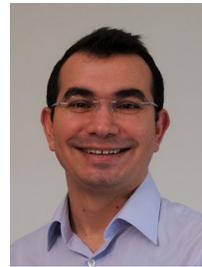

**Mustafa Ridvan Cantas** earned his M.S. and B.S degrees in Electrical and Electronics Engineering from Bilkent University, Ankara, Turkey, in 2012, and Yeditepe University, Istanbul, Turkey, in 2009, respectively. He is currently working toward his Ph.D. degree in the Automated Driving Lab, Department of Electrical and Computer Engineering, The Ohio State University, Columbus, OH, USA. His research interests include intelligent transportation systems, advanced driver assistance systems, connected and autonomous vehicles. He is a Member of IEEE.

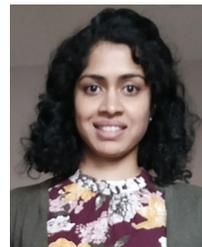

**Arpita Chand** earned her M.S. in Electrical Engineering from the State University of New York at Buffalo in 2016, Buffalo, NY, USA. She is currently working as an Advanced Connectivity Technology Engineer at Ford Motor Company in Michigan. Her research interests include wireless technologies in connected vehicles, typically V2X communication, multi-access edge computing for vehicle and smart city applications, and machine learning algorithms for driver assistance technologies.

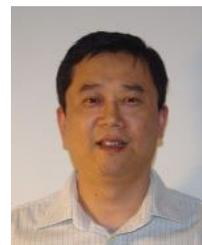

**Hao Zhang** earned his M.S. in Electrical Engineering and Computer Science from Case Western Reserve University, Cleveland, OH in 2000. He is currently a production design Engineer working in Research and Advanced Engineering in Ford Motor Company, his main research areas is collision avoidance using Driver Assistant Technologies and Vehicle Connectivity Technology.

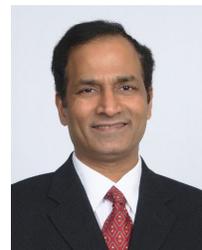

**Gopi C. Surnilla** earned his M.S.ME and M.B.A from University of Akron, OH. He is a Technical Leader for Control & Technology Innovation and Implementation at Ford Motor Company. He has been working at Ford for over 26 years. His expertise is in controls research in the areas Driver Assist Technologies (DAT), Connectivity, Data Analytics & Powertrain. Currently, he is researching C-V2X technology & applications for safety, efficiency and traffic flow optimization. He is a strong innovator in the automotive field and has over 500 patents to his credit. He also leads development of innovation processes & skills.

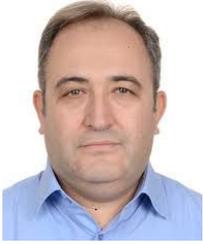 **Levent Guvenc** earned his Ph.D. degree in mechanical engineering from the Ohio State University, Columbus, OH, USA, in 1992. He is currently a professor in mechanical and aerospace engineering with the Ohio State University with a joint appointment at the electrical and computer engineering department. His research interests include Automotive Control, ADAS, Connected Vehicles and Autonomous Driving. He is a Member of IEEE.